\documentclass{article}
\usepackage{spconf,amsmath,graphicx}
\usepackage{graphicx}
\usepackage{times}
\usepackage{url}
\usepackage[utf8]{inputenc}
\usepackage[small]{caption}
\usepackage{graphicx}
\usepackage{amsmath}
\usepackage{amsthm}
\usepackage{booktabs}
\usepackage{algorithm}
\usepackage{algorithmic}
\usepackage{amssymb}
\usepackage{booktabs}
\usepackage{multirow}
\usepackage{utfsym}
\usepackage{amssymb}
\usepackage{amsmath}

\usepackage{pdfpages}
\usepackage{xcolor}
\usepackage{colortbl}

\newcommand{\revise}[1]{}
\newcommand{\secondrevise}[1]{}

\usepackage[colorlinks,
            linkcolor=blue,
            anchorcolor=blue, 
            citecolor=blue,
            ]{hyperref}
\title{Refining Open-Vocabulary Semantic Segmentation via Regional Semantics and Visual Prototypes}

\name{Aijing Yu$^{1,2}$, Zhengbo Wang$^{4,5}$, Lijun Sheng$^{4,5}$, Jian Liang$^{3,4}$, Xiao-Yu Zhang$^{1,2}$\thanks{Xiao-Yu Zhang is the corresponding author (zhangxiaoyu@iie.ac.cn).}}

\address{
$^{1}$ School of Cyber Security, University of Chinese Academy of Sciences\\
$^{2}$ Institute of Information Engineering, Chinese Academy of Sciences\\
$^{3}$ School of Artificial Intelligence, University of Chinese Academy of Sciences \\
$^{4}$ NLPR \& MAIS, Institute of Automation, Chinese Academy of Sciences \\
$^{5}$ University of Science and Technology of China \\
}
\begin{document}
\ninept
\maketitle
\begin{abstract}
Contrastive Language-Image Pre-training (CLIP) has shown superior performance in the image-level open-vocabulary classification task, benefiting from its powerful alignment of visual features and text embeddings.
However, dense prediction tasks require models to perform pixel-level predictions, necessitating modifications to CLIP for region-text alignment.
In this paper, we adjust the attention layers within the vanilla visual encoder of CLIP.
With a novel attention calculation method for selective attention features, the model retains more regional semantic information.
To further mitigate the negative impact of the modality gap, we introduce additional visual prototypes to refine the original predictor solely based on text embeddings.
Specifically, patch features with confident predictions are selected to form an additional auxiliary memory-based classifier for more accurate prediction.
We further integrate an optimizable feature extractor with an exponential moving average updating strategy to enhance prediction quality.
Extensive experiments on $5$ semantic segmentation datasets demonstrate the effectiveness of our method.
\end{abstract}
\section{Introduction}
\label{sec:intro}

Semantic segmentation~\cite{long2015fully,wang2024curriculum} aims to classify an image at the pixel level, applying segmentation masks to separate diverse objects in the input.
Conventional semantic segmentation methods follow a supervised paradigm, relying on manually labeled category annotations.
To address the limitations of predefined categories, the Open-Vocabulary Semantic Segmentation (OVSS) task has been proposed and is receiving widespread attention.
OVSS aims to effectively handle entirely new categories during inference, which may not have appeared during training.
Recently, visual language models~\cite{radford2021learning,jia2021scaling} have shown remarkable generalization performance in image-level classification tasks~\cite{liang2024realistic,dong2026adapting}.
Inspired by this, some existing OVSS methods explore the potential of Contrastive Language-Image Pre-training (CLIP)~\cite{radford2021learning} in pixel-level classification tasks.
Most of them~\cite{jia2021scaling,zhang2024uncovering,qin2023freeseg} follow a semi-supervised paradigm, utilizing both image-text pairs and additional dense annotations.
Typically, these methods train a dense predictor with segmentation masks and expand target categories using image-text pairs~\cite{ghiasi2022scaling,liang2023open}. 
Furthermore, recent works improve feature extraction with additional image-text pairs~\cite{cha2023learning}, clustering features~\cite{liu2022open,xu2022groupvit}, modifying structures~\cite{zhou2022extract,shin2022reco}, aiming to convert image-level visual-text alignment to pixel-level visual-text alignment.
MaskCLIP~\cite{zhou2022extract} revisits the global pooling layer of the visual encoder in the CLIP model, arguing that its \textit{value} features already encompass rich pixel-level semantic information. 
Consequently, it directly calculates the similarity between the \textit{value} features of the last attention module and the text embeddings to obtain accurate segmentation masks.
However, the segmentation masks generated by MaskCLIP still exhibit noise.
Recent work~\cite{li2023clip} suggests that while using the final \textit{value} projection for prediction, the \textit{key} and \textit{query} projection in the attention module may concatenate features that do not belong to the same semantic category, thereby diminishing the correlation between pixels in neighboring regions.
Additionally, existing methods remain constrained by the inherent vision-text modality gap of the CLIP.

To address the challenging OVSS task, we propose an annotation-free method to obtain high-quality segmentation masks in this work.
We begin by introducing a regional semantic preservation module to refine pixel-level visual features.
In this module, we modify the attention layer of the original CLIP image encoder, selectively reusing features that contain rich pixel-level semantic information.
We design a novel attention calculation method for these features, enhancing regional semantic coherence by calculating the similarity of adjacent pixel-level features.
To preserve local information and maintain the integrity of feature extraction, we modify only the final attention layer in pre-trained CLIP.

To mitigate the vision-language modality gap in CLIP, we further leverage additional visual prototypes of the entire unlabeled set to refine the dense predictions.
For each semantic category, we select representative pixel-level features to form a visual feature memory module.
Serving as an auxiliary dense predictor, the memory module computes the similarity of its vision embedding with the visual feature of the input image.
The dense predictions from the auxiliary predictor are combined with the outcomes by text embeddings to construct the final segmentation results.
The core idea of this design is to use the intra-domain similarity of the vision embedding and the input image to alleviate the vision-language modality gap inherent in CLIP. 
Additionally, we incorporate an additional feature extractor specifically tailored for semantic segmentation, optimizing it to enhance prediction performance further.
The optimization is entirely supervised using model-generated masks of unlabeled samples, without relying on any annotated information during the process.
To improve generalization and stability, we adopt an Exponential Moving Average (EMA) strategy to maintain a smoothly updated version of the feature extractor.

In summary, our contributions are as follows:
\begin{itemize}
\item We propose a novel attention modification method based on vanilla CLIP to preserve regional semantics.
We intuitively modify the final attention layer of CLIP's image encoder, devising a specific attention calculation method to obtain pixel-level visual features with greater local semantic consistency.
\item We introduce a prediction enhancement strategy with visual prototypes. 
A visual feature memory module is designed to select representative visual prototypes for each category. 
Serving as an auxiliary dense predictor, this module mitigates the modality gap of text embeddings and images.
\item We achieve state-of-the-art annotation-free semantic segmentation performance compared to existing methods.
\end{itemize}

\begin{figure}[!t]
\centering
\includegraphics[width=0.48\textwidth, height=0.32\textwidth]{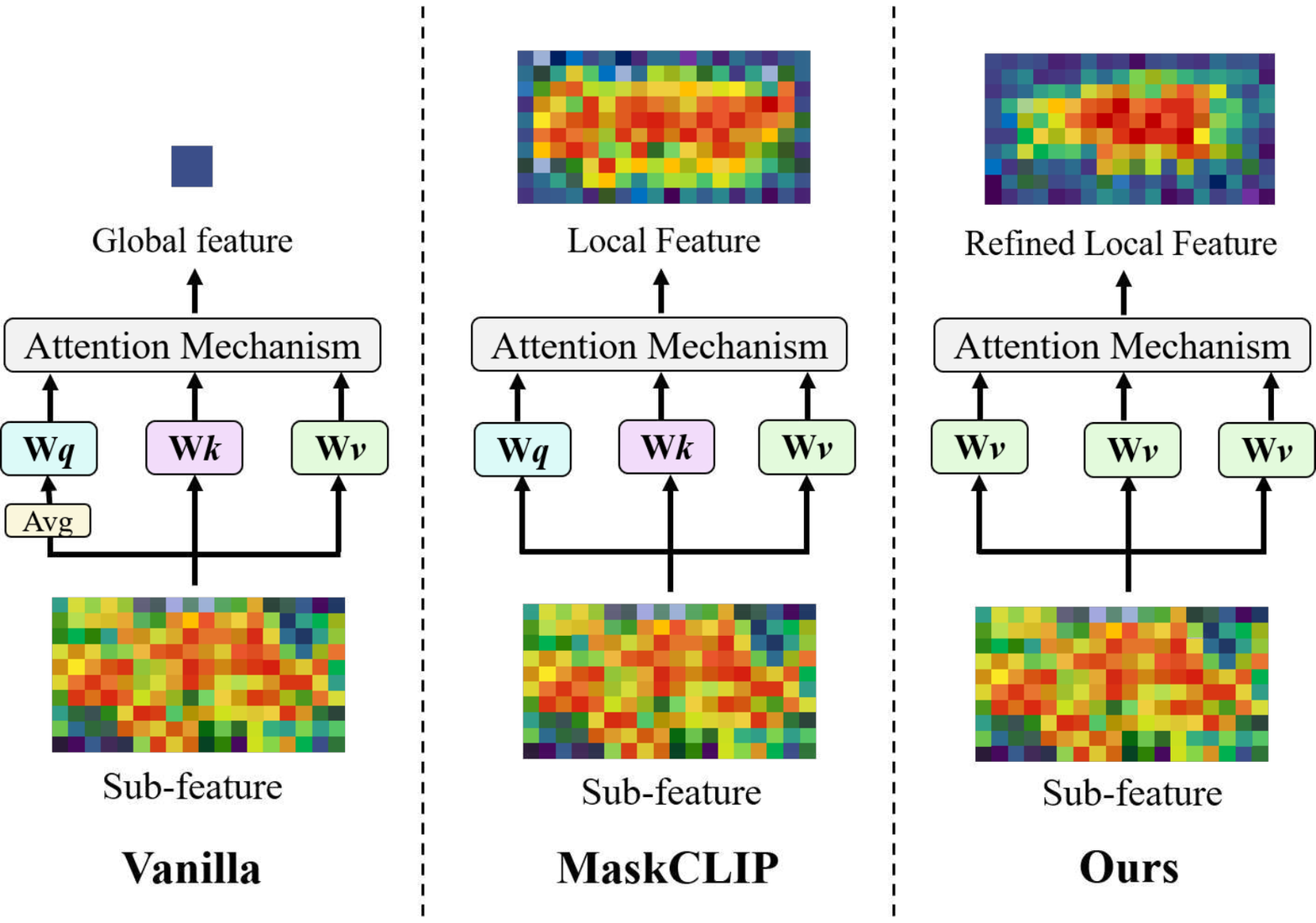}
\caption{The diagram of the Regional Semantic Preservation Module. We discard \textit{Wq} and \textit{Wk}, using only the \textit{value feature} output by \textit{Wv} for self-attention calculation. This mechanism yields dense features that better capture regional semantic information.}
\label{fig1}
\end{figure}

\section{Methodology}
\label{method}
\subsection{Preliminaries}
\label{3.2}

\begin{figure*}[!t]
\centering
\includegraphics[width=0.78\textwidth]{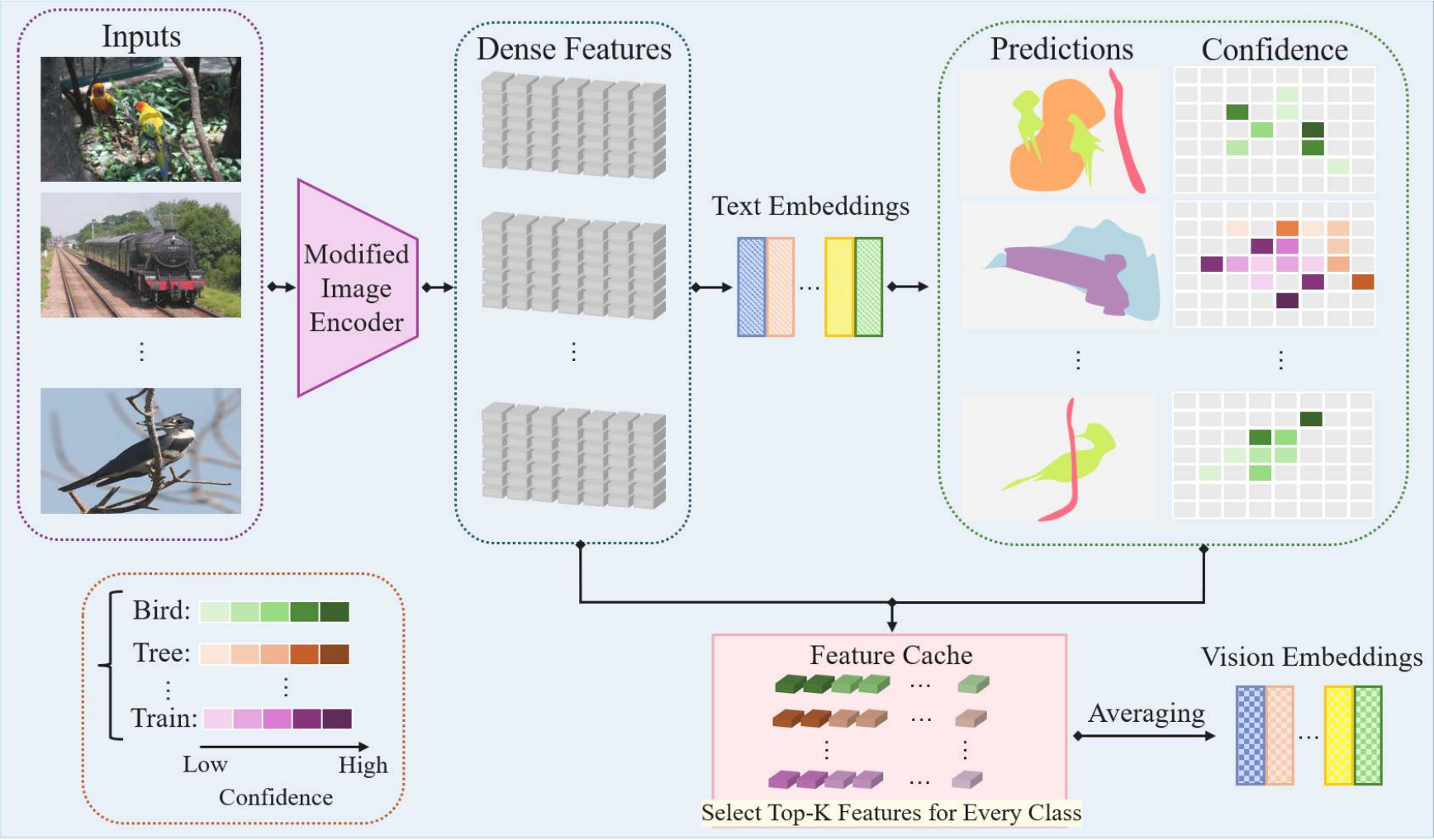}
\caption{The pipeline of Visual Feature Memory Module. Given the inputs, we employ the regional semantic preservation module in Section~\ref{3.3} to extract the dense features. Then, the text embeddings of CLIP output the class prediction and the corresponding probability confidence. The feature cache is used to select the visual features with Top-K confidence, which are averaged and set as vision embeddings. }
\label{fig2}
\end{figure*}

{\bf Contrastive Language-Image Pre-Training (CLIP)} model achieves cross-modal understanding by jointly training a text encoder $\mathcal{T}$ and an image encoder $\mathcal{I}$ with abundant image-text pairs.
For a K-class classification task, each class has a text description $d_j$, such as the category name with a unified prompt prefix.
Consider a test image $x_i$, CLIP model first utilizes the image encoder to extract its visual feature $f_i = \mathcal{I}(x_i)$.
To aggregate the sub-features of each spatial position in image feature extraction, image encoder $\mathcal{I}(\cdot)$ contains an attention-based pooling layer.
It treats the averaged sub-feature as the query feature and all sub-features as key-value features, the self-attention mechanism is computed as,
\begin{equation}
  \operatorname{Attn}_{q k} = \sum_i \operatorname{softmax}\left(\frac{\bar{q} k_i^{\top}}{\sqrt{C}}\right) v_i,
  \label{eq:clip_attn}
\end{equation}
where the $k_i$ and $v_i$ are key and value features extracted by the linear embedding layer at each spatial position $i$, and $\bar{q}$ is the average of all query features.
The text description is fed into the text encoder to obtain textual feature $t_j = \mathcal{T}(d_j)$.
The final prediction probability is calculated as follows:
\begin{equation}
  p(y=j | x_i) = \frac{\exp( \cos(t_j, f_i)/\tau) } {\sum_{c=1}^K \exp( \cos(t_c, f_i) /\tau)},
  \label{eq:clip_zero_shot}
\end{equation}
where $\cos$ represents the cosine similarity and $\tau$ is the temperature.

\begin{figure}[!t]
\centering
\includegraphics[width=0.45\textwidth, height=0.25\textwidth]{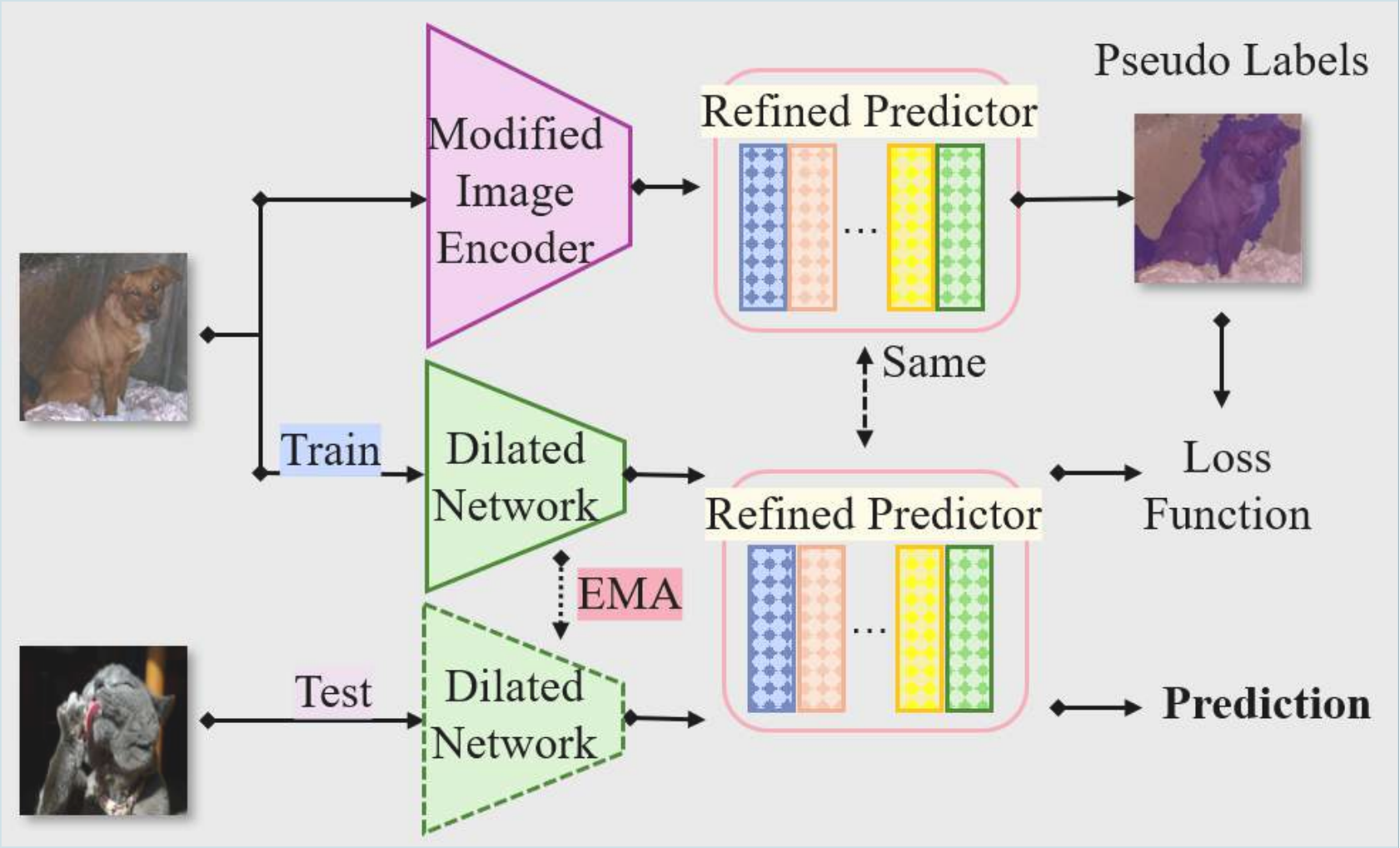}
\caption{Overview of our method. The input image is encoded by the regional semantic preservation module in Section~\ref{3.3}, followed by the refined predictor in Section~\ref{3.4} for semantic segmentation.}
\label{fig3}
\end{figure}

{\bf MaskCLIP} presents the hypothesis that the computed $\Phi \left( {v_i}\right)$ at each spatial position already captures rich local semantic responses, well aligned with the text embeddings of the CLIP.
Therefore, 
MaskCLIP directly inputs ${v_i}$ into the text-embedding based predictor, discarding the keys and queries, and achieves amazing performance improvements on annotation-free segmentation tasks.

\subsection{Regional Semantic Preservation Module}
% (RSPM)
\label{3.3}
Although $\Phi \left( {v_i}\right)$ contains abundant per-pixel semantic information, it overlooks the semantic correlation among neighboring regions. Additionally, the \textit{q-k mapping} in the self-attention module encourages the model to focus on the global information of images, which may not be suitable for dense prediction tasks. 
To address this issue, we propose the Regional Semantic Preservation Module, aimed at guiding the model to focus more on regional semantic information.
Specifically, as demonstrated in Figure~\ref{fig1}, we introduce a novel self-attention calculation method to promote the consistency of visual features in adjacent spatial regions. 
We modify the last attention layer in the pre-trained CLIP and employ a \textit{v-v mapping} to replace the original \textit{q-k mapping}, which is defined as:
\begin{equation}
  \operatorname{Attn}_{v v} = \sum_i \operatorname{softmax}\left(\frac{\bar{v} v_i^{\top}}{\sqrt{C}}\right) v_i.
  \label{eq:eq3}
\end{equation}
This intuitive modification emphasizes the similarity of adjacent pixel-level features and enhances the clustering of nearby features, aggregating semantically related areas while mitigating attention to semantically irrelevant regions.
We only revise the attention layer within the final self-attention block, aiming to enhance the prediction performance while preserving feature extraction functionality with minimal adjustments.

\subsection{Prediction Enhancement with Visual Prototypes}
\label{3.4}
In this section, we present a Prediction Enhancement strategy with Visual Prototypes, aiming to refine the classification outputs from text embeddings.
As shown in Figure~\ref{fig2}, we employ a feature cache module to store the most representative visual features for each semantic category.
The vision embeddings in the memory module are employed as an additional predictor, to align features within the visual modality and, to some extent, alleviate the modality gap between text and vision embeddings.
The predicted results of vision embeddings and text embeddings are combined in a residual way, which reconciles the final segmentation results while retaining the rich semantic knowledge in the text embeddings of CLIP. 

Specifically, for each image $X_{i}\in \mathbb{R} ^{H\times W\times C}$ in the target dataset $\mathcal{D}$, we first input them into MaskCLIP for preliminary dense prediction.
we extract pixel-level visual features for each image using the modified image encoder $\Psi _{\mathcal{I} } $ of MaskCLIP in Section~\ref{3.3}: 
\begin{equation}
  \left \{ F_1^{image},\dots ,F_N^{image} \right \} =\left \{ \Psi _{\mathcal{I}}\left ( x_1 \right ) ,\dots, \Psi _{\mathcal{I}}\left ( x_N \right )\right \}.
  \label{eq:eq4}
\end{equation}
Meanwhile, we use $\left \{ F_c^{text}\right \}_{c=1}^{C} $ and $\mathit{C} _{\mathfrak{t} } \left ( \cdot  \right ) $ to denote text embeddings for each semantic category and the text-embedding based predictor, respectively.
For an image $X_{i}$, the category prediction probability at each spatial location $s$ can be obtained by Calculating the similarity between its features $F_i^{image}$ and text embeddings, which can be abbreviated as:
\begin{equation}
  \mathit{p_s} = \mathit{C} _{\mathfrak{t} } \left ( F_n^{image}\right ).
  \label{eq:eq5}
\end{equation}
We select the visual features $\left \{ F_{{c}_{1}}^{pixel},\dots, F_{{c}_{K}}^{pixel}\right \}$ corresponding to pixels with \textit{top-K} category prediction confidence for each target category $c$.
The visual feature prototype of each target category is obtained by averaging the \textit{top-K} visual features as:
\begin{equation}
  \overline{F}_{c} ^{pixel}  =\frac{\sum_{k=1}^{K} F_{{c}_{k}}^{pixel}}{K}.
  \label{eq:eq6}
\end{equation}
We normalize the visual feature prototype into the same form as the text embeddings, which is named pixel embedding $\left \{ F_c^{pixel}\right \}_{c=1}^{C} $, thereby obtaining the vision-embedding based predictor $\mathit{C} _{\mathfrak{p} } \left ( \cdot  \right ) $.
The final semantic result is then calculated as:
\begin{equation}
  \mathit{p} = \mathit{C} _{\mathfrak{t} } \left ( F^{image}\right ) + \lambda \mathit{C} _{\mathfrak{p} } \left ( F^{image}\right ),
  \label{eq:eq7}
\end{equation}
where $\lambda$ is the residual ratio.
In this manner, the refined predictor encompasses both prediction outcomes of the visual modality and the original text-image alignment results. 

\subsection{Self-training}
\label{3.5}
To fully harness the capabilities of the segmentation network, we replace the vision encoder of CLIP with an optimized auxiliary feature extractor customized for semantic segmentation.
We utilize the segmentation outcomes generated by the refined predictor in Section~\ref{3.4} as pseudo-labels for optimization.
To express concisely, the revised predictor is ensembled by $\mathit{C} _{\mathfrak{t} } \left ( \cdot  \right ) $ and $\mathit{C} _{\mathfrak{p} } \left ( \cdot  \right ) $, which is expressed as $\mathit{C} _{\mathfrak{r} } \left ( \cdot  \right ) $.
We replace the original classifiers with $\mathit{C} _{\mathfrak{r} } \left ( \cdot  \right ) $, to produce open-vocabulary semantic segmentation.
Additionally, we employ the Exponential Moving Average (EMA) in the inference stage to smooth the learned parameters of the target feature extractor, $\text{EMA}_t=\beta \cdot X_t+(1-\beta) \cdot \text{EMA}_{t-1}$.
We follow the official settings ($\beta=0.9999$) for model inference. 

\section{Experiments}
\label{exper}
\subsection{Implementation Details}
\label{4.1}
We use $5$ standard segmentation benchmarks to evaluate our method, namely ADE20K~\cite{zhou2019semantic}, Cityscapes~\cite{cordts2016cityscapes}, COCO Stuff164k~\cite{caesar2018coco}, PASCAL Context~\cite{mottaghi2014role}, and VOC12AUG~\cite{everingham2010pascal}.
This work considers two publicly available visual backbones, namely CLIP-ResNet-50~\cite{radford2021learning} and CLIP-ViT-B/16~\cite{radford2021learning} models.
For CLIP-ResNet-50, the modification involves rewriting the final global pooling layer, replacing both \textit{q-projection} and \textit{k-projection} with \textit{v-projection}.
For CLIP-ViT-B/16, we directly reuse the output \textit{v} of the last attention block to conduct additional self-attention calculations.
In the training-free prediction, the refined predictor proposed in Section~\ref{3.5} directly outputs segmentation results.
The settings of \textit{top-K} $\&$ $\lambda$ are shown in Table~\ref{tabh}.
We use 85 predefined prompt templates to encode corresponding text embeddings and average them for each category as in ViLD~\cite{gu2021open}.
Following MaskCLIP+, we employ DeepLabv2-ResNet101~\cite{chen2018encoder} which is initialized with the ImageNet~\cite{deng2009imagenet} pre-trained weights as the target model.
Our method is implemented based on the MMSegmentation codebase and executed on 4 Nvidia 4090 GPUs.

\begin{table}[!t]
  \centering
  \renewcommand\arraystretch{0.8}
  \caption{
  Hyper-parameter settings in our method across different datasets.
  $^\dagger$ denotes hyper-parameters in the distillation process.
  }
  \label{tabh}
  \resizebox{0.45\textwidth}{!}{
    \begin{tabular}{lccccc}
    \toprule
        Datasets  & ADE  & City & COCO & PC  & VOC \\
    \midrule
    Top-K & 512   & 128   & 4,096  & 1,024  & 128 \\
    $\lambda$ & 0.5   & 0.05  & 0.5   & 0.5   & 0.5 \\
    Training Iterations$^\dagger$ & 8,000  & 4,000  & 8,000 & 4,000  & 2,000 \\
    \bottomrule
    \end{tabular}%
 }
  \label{tab:addlabel}%
\end{table}%

\begin{table}[]
\small
\centering
\renewcommand\arraystretch{0.9}
\caption{Quantitative results (mIoU, $\%$) comparison on $5$ semantic segmentation benchmarks. MaskCLIP$^{\dagger}$ means their baseline without refinement. We highlight the  \textbf{best} and the \underline{second-best} results.}
\label{tab1}
\resizebox{0.45\textwidth}{!}{
\begin{tabular}{lcccccc}
\toprule
Method        & ADE & City & COCO & PC & VOC & \textbf{Avg.} \\ \hline
CLIP~\cite{radford2021learning} &3.1 &6.5 &5.7 &11.1 &49.4 &14.1 \\ \hline
MaskCLIP+$^{\dagger}$~\cite{zhou2022extract} &10.8 &21.6 &14.7 &23.3 &53.7 &23.7 \\
MaskCLIP+~\cite{zhou2022extract} &11.9 &24.9 &16.7 &26.4 &74.9 &30.3 \\ \hline
GroupViT~\cite{xu2022groupvit} & 10.4 & 18.5 & 15.3 & 23.4 & \textbf{79.7} & 30.7 \\
ReCLIP~\cite{wang2024learn}&14.3&19.9&20.2&33.7&75.8&32.8\\
ReCo~\cite{shin2022reco} & 11.2 & 21.6 & 14.8 & 22.3 & 57.7 & 23.5 \\
TCL~\cite{cha2023learning} & 14.9 & 23.5 & 19.6 & 30.3 & \underline{77.5} & \underline{33.9} \\
CLIP Surgery~\cite{li2023clip}  & 12.9 & \underline{31.4} &\textbf{21.9} & - & - & - \\
CoCu~\cite{xing2023rewrite}&11.1&15.0&13.6&-&-&-\\
GEM~\cite{bousselham2024grounding} & \underline{15.9} & - & - & \underline{32.6} & - & - \\ \hline
\multirow{2}{*}{\textbf{Ours}} & \textbf{17.2} & \textbf{32.5} & \underline{20.3} & \textbf{33.7} & 75.7 & \textbf{35.9} \\
 &(\textbf{\textcolor{magenta}{+1.3}}) &(\textbf{\textcolor{magenta}{+1.1}}) & &(\textbf{\textcolor{magenta}{+1.1}})& &(\textbf{\textcolor{magenta}{+2.0}})\\
\bottomrule
\end{tabular}
}
% \vspace{0pt}
\end{table}

\begin{table}[!t]
% \small
\centering
\caption{Quantitative results (mIoU, $\%$) under CLIP-ResNet-50. Ours with $\diamondsuit$ symbol denotes the training-free setting.}
\label{tab2}
\renewcommand\arraystretch{0.9}
\resizebox{0.45\textwidth}{!}{
\begin{tabular}{lcccccc}
\toprule
Method        & ADE & City & COCO & PC & VOC & \textbf{Avg.} \\ \hline
MaskCLIP &8.3 &18.8 &10.2 &18.5 &41.5 &19.5 \\ 
\textbf{Ours$^{\diamondsuit}$} & 11.5 & 22.1 & 15.0 & 24.4 & 59.0 & 26.4 \\ \hline
MaskCLIP+ &10.7 &23.4 &13.6 &24.2 &57.3 &25.8 \\ 
\textbf{Ours} & 13.0 & 26.9 & 16.4 & 28.0 & 70.2 & 30.9 \\
\bottomrule
\end{tabular}
}
\end{table}

\begin{figure}[!t]
\centering
\includegraphics[width=0.48\textwidth]{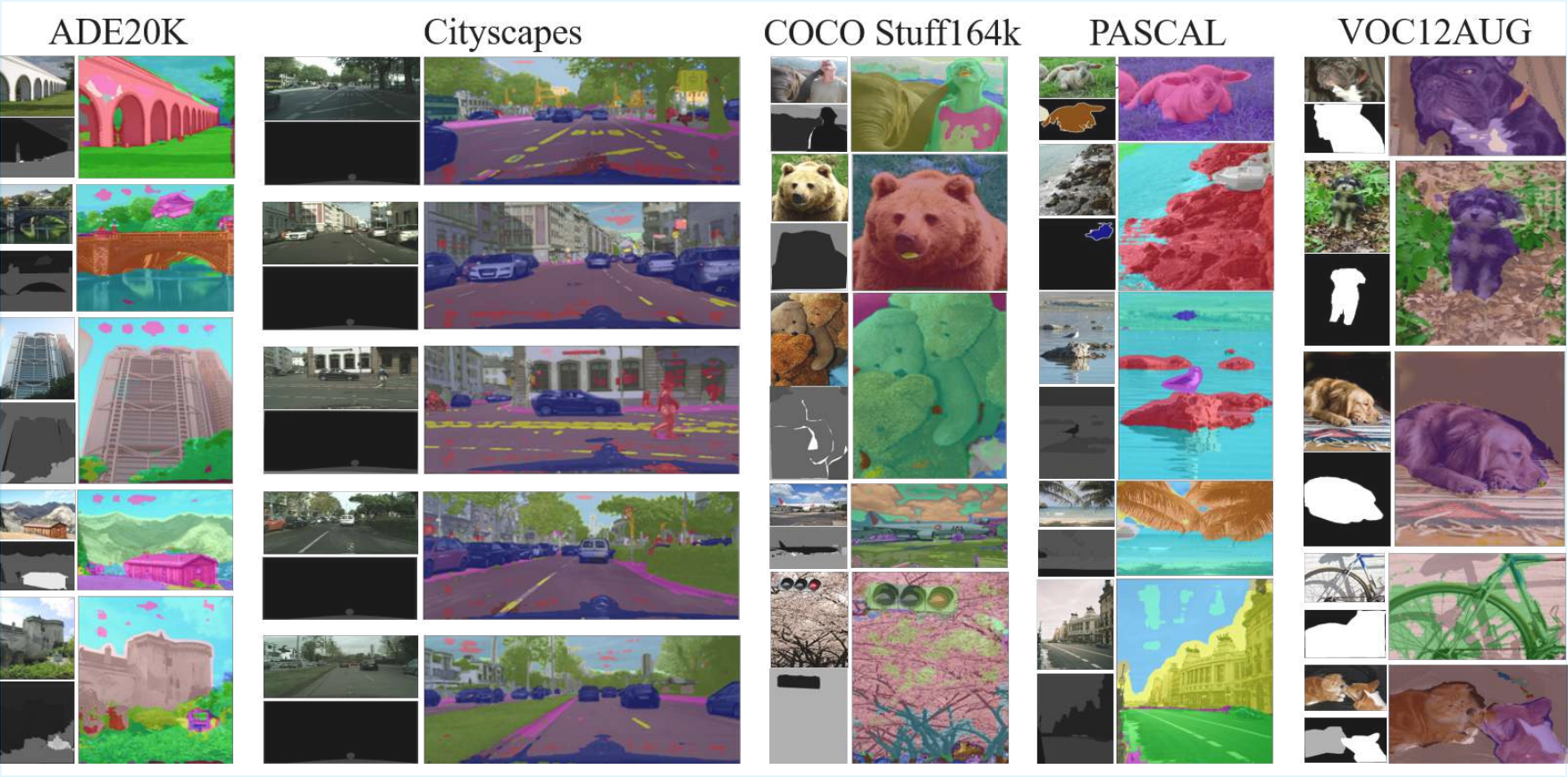}
\caption{Qualitative results on all the $5$ datasets. Each $3$ closely arranged images form a group, in which the upper left corner, lower left corner, and right image are respectively the input image, ground truth, and the visualization result of the proposed method.}
\label{Qual}
\end{figure}

\begin{table}[!t]
% \small
\centering
\caption{Ablation results (mIoU, $\%$) of each component in our method. A., V., and E. denote the refined self-attention in Section~\ref{3.3}, vision embeddings saved by the visual feature memory module in Section~\ref{3.4}, and the optimization strategy in Section~\ref{3.5}, respectively.}
\label{tab3}
\renewcommand\arraystretch{0.9}

\resizebox{0.48\textwidth}{!}{\begin{tabular}{clcccccc}
\toprule
Self-train & \multicolumn{1}{c}{Method} & ADE & City & COCO & PC & VOC & \textbf{Avg.} \\ \hline 
\multirow{3}{*}{$\usym{1F5F4}$} & MaskCLIP  & 9.5 & 19.8 & 12.5 & 21.7 & 49.5 & 22.6 \\ 
  & + A. & 10.0   & 20.6       & 13.2 & 22.7           & 51.4     & 23.6  \\
  & + A. + V. & \textbf{12.6}   & \textbf{20.7}       & \textbf{15.3}           & \textbf{25.1}           & \textbf{56.2}     & \textbf{26.0} \\ \hline
\multirow{4}{*}{$\usym{2714}$} & MaskCLIP+  & 11.9   & 24.9   & 16.7 & 26.4  & 74.9     & 31.0 \\
  & + A. & 15.2   & 32.3       & 19.1           & 32.4           & 75.3     & 34.9  \\
  & + A. + V. & 17.1   & 32.1       & 20.2           & 32.6           & 75.6     & 35.5  \\
  & + A. + V. + E.  & \textbf{17.2}   & \textbf{32.5}       & \textbf{20.3}           & \textbf{33.7}           & \textbf{75.7}     & \textbf{35.9} \\
  \bottomrule
\end{tabular}

}
\end{table}

\subsection{Main Results}
\label{4.2}
\textbf{Comparison to Open-vocabulary semantic segmentation Methods.} 
In Table~\ref{tab1}, we present the comparison results (mIoU, $\%$) of our method with other existing open-vocabulary semantic segmentation models, including MaskCLIP~\cite{zhou2022extract}, GroupViT~\cite{xu2022groupvit}, ReCo~\cite{shin2022reco}, TCL~\cite{cha2023learning}, CLIP Surgery~\cite{li2023clip}, and GEM~\cite{bousselham2024grounding}.
For fairness in comparison, we do not include the additional PAMR post-processing layer in any of the methods.
As demonstrated in In Table~\ref{tab1}, our method achieves state-of-the-art performance on ADE20K(17.2$\%$), Cityscapes(32.5$\%$), and PASCAL Context(33.7$\%$).
Besides, the average performance of ours reaches 35.9$\%$, which is also the highest compared to other methods.

\textbf{Experiments on Other Visual Backbone.} 
Our approach is also applicable to CLIP-ResNet-50.
In Table~\ref{tab2}, we present the quantitative evaluation across $5$ datasets. 
Both the training-free prediction and self-training setting are considered in the evaluation.
The results demonstrate that our method can still enhance dense prediction performance on CLIP-ResNet-50 to a certain extent.

\textbf{Qualitative Results.} 
In Figure~\ref{Qual}, we showcase the qualitative outcomes of our method across $5$ diverse datasets. 
Each column corresponds to images sourced from the same dataset.
In each row of Figure~\ref{Qual}, for each dataset, $3$ images form a combination.
Among them, the top-left image represents the input image, the bottom-left image depicts the ground truth, and the image on the right showcases the dense prediction results produced by our method.
The experimental results demonstrate our method's ability to accurately delineate object boundaries, ensuring precise localization, and maintaining a nuanced understanding of local context. 

\textbf{Ablation Study}
\label{4.3}
We analyze the effectiveness of each component in the proposed method.
Through a series of ablation experiments conducted across all $5$ datasets utilizing the CLIP-ViT-B/16 backbone, we meticulously assess the impact of various elements on overall performance. 
In Table~\ref{tab3}, we present the outcomes obtained from both training-free prediction and self-training.
The experimental results indicate that our proposed components yield significant improvements across all datasets.

\section{Conclusion}
\label{concl}
In this work, we focus on open-vocabulary semantic segmentation with an annotation-free setting. 
In this task, dense features generated by CLIP's visual encoder are utilized to align with text embeddings for dense prediction.
We suggest that the inherent attention mechanism of the image encoder in CLIP is not entirely suitable for semantic segmentation, as the visual features tend to focus on the global information of the image rather than pixel-level semantics.
We propose a novel attention computation approach that encourages the image encoder to preserve semantic consistency among features from adjacent regions to a greater extent.
Additionally, we introduce a feature memory module that stores visual feature prototypes as an additional predictor, collaborating with text embeddings for dense prediction.
Our approach achieves state-of-the-art performance across multiple benchmarks, demonstrating the superiority of the proposed components.

\section{acknowledgments}
\label{ack}
This work was supported by the National Natural Science Foundation of China (NSFC) (Grant 62376265).
It was also supported by the National Natural Science Foundation of China under Grants (62276256, U2441251).

\section{ethical standards statement}
\label{eth}
This is a numerical simulation study for which no ethical approval was required.

\bibliographystyle{IEEEbib}
\bibliography{refs}

@String(CVPR = {Proc. CVPR})

@String(ECCV = {Proc. ECCV})

@String(ICML   = {Proc. ICML})

@String(ICLR   = {Proc. ICLR})

@inproceedings{long2015fully,
  title={Fully convolutional networks for semantic segmentation},
  author={Long, Jonathan and Shelhamer, Evan and Darrell, Trevor},
  booktitle={Proc. CVPR},
  pages={3431--3440},
  year={2015}
}

@inproceedings{radford2021learning,
  title={Learning transferable visual models from natural language supervision},
  author={Radford, Alec and Kim, Jong Wook and Hallacy, Chris and Ramesh, Aditya and Goh, Gabriel and Agarwal, Sandhini and Sastry, Girish and Askell, Amanda and Mishkin, Pamela and Clark, Jack and others},
  booktitle={Proc. ICML},
  pages={8748--8763},
  year={2021},
}

@inproceedings{jia2021scaling,
  title={Scaling up visual and vision-language representation learning with noisy text supervision},
  author={Jia, Chao and Yang, Yinfei and Xia, Ye and Chen, Yi-Ting and Parekh, Zarana and Pham, Hieu and Le, Quoc and Sung, Yun-Hsuan and Li, Zhen and Duerig, Tom},
  booktitle={Proc. ICML},
  pages={4904--4916},
  year={2021},
}

@inproceedings{zhang2024uncovering,
  title={Uncovering prototypical knowledge for weakly open-vocabulary semantic segmentation},
  author={Zhang, Fei and Zhou, Tianfei and Li, Boyang and He, Hao and Ma, Chaofan and Zhang, Tianjiao and Yao, Jiangchao and Zhang, Ya and Wang, Yanfeng},
  booktitle={Proc. NeurIPS},
  volume={36},
  year={2023}
}

@inproceedings{qin2023freeseg,
  title={Freeseg: Unified, universal and open-vocabulary image segmentation},
  author={Qin, Jie and Wu, Jie and Yan, Pengxiang and Li, Ming and Yuxi, Ren and Xiao, Xuefeng and Wang, Yitong and Wang, Rui and Wen, Shilei and Pan, Xin and others},
  booktitle={Proc. CVPR},
  pages={19446--19455},
  year={2023}
}

@inproceedings{ghiasi2022scaling,
  title={Scaling open-vocabulary image segmentation with image-level labels},
  author={Ghiasi, Golnaz and Gu, Xiuye and Cui, Yin and Lin, Tsung-Yi},
  booktitle={Proc. ECCV},
  pages={540--557},
  year={2022},
  organization={Springer}
}

@inproceedings{liang2023open,
  title={Open-vocabulary semantic segmentation with mask-adapted clip},
  author={Liang, Feng and Wu, Bichen and Dai, Xiaoliang and Li, Kunpeng and Zhao, Yinan and Zhang, Hang and Zhang, Peizhao and Vajda, Peter and Marculescu, Diana},
  booktitle={Proc. CVPR},
  pages={7061--7070},
  year={2023}
}

@inproceedings{cha2023learning,
  title={Learning to generate text-grounded mask for open-world semantic segmentation from only image-text pairs},
  author={Cha, Junbum and Mun, Jonghwan and Roh, Byungseok},
  booktitle={Proc. CVPR},
  pages={11165--11174},
  year={2023}
}

@inproceedings{liu2022open,
  title={Open-world semantic segmentation via contrasting and clustering vision-language embedding},
  author={Liu, Quande and Wen, Youpeng and Han, Jianhua and Xu, Chunjing and Xu, Hang and Liang, Xiaodan},
  booktitle={Proc. ECCV},
  pages={275--292},
  year={2022},
  organization={Springer}
}

@inproceedings{xu2022groupvit,
  title={Groupvit: Semantic segmentation emerges from text supervision},
  author={Xu, Jiarui and De Mello, Shalini and Liu, Sifei and Byeon, Wonmin and Breuel, Thomas and Kautz, Jan and Wang, Xiaolong},
  booktitle={Proc. CVPR},
  pages={18134--18144},
  year={2022}
}

@inproceedings{zhou2022extract,
  title={Extract free dense labels from clip},
  author={Zhou, Chong and Loy, Chen Change and Dai, Bo},
  booktitle={Proc. ECCV},
  pages={696--712},
  year={2022},
  organization={Springer}
}

@inproceedings{shin2022reco,
  title={Reco: Retrieve and co-segment for zero-shot transfer},
  author={Shin, Gyungin and Xie, Weidi and Albanie, Samuel},
  booktitle={Proc. NeurIPS},
  volume={35},
  pages={33754--33767},
  year={2022}
}

@article{li2023clip,
  title={Clip surgery for better explainability with enhancement in open-vocabulary tasks},
  author={Li, Yi and Wang, Hualiang and Duan, Yiqun and Li, Xiaomeng},
  journal={arXiv preprint arXiv:2304.05653},
  year={2023}
}

@inproceedings{chen2018encoder,
  title={Encoder-decoder with atrous separable convolution for semantic image segmentation},
  author={Chen, Liang-Chieh and Zhu, Yukun and Papandreou, George and Schroff, Florian and Adam, Hartwig},
  booktitle={Proc. ECCV},
  pages={801--818},
  year={2018}
}

@article{zhou2019semantic,
  title={Semantic understanding of scenes through the ade20k dataset},
  author={Zhou, Bolei and Zhao, Hang and Puig, Xavier and Xiao, Tete and Fidler, Sanja and Barriuso, Adela and Torralba, Antonio},
  journal={International Journal of Computer Vision},
  volume={127},
  pages={302--321},
  year={2019},
  publisher={Springer}
}

@inproceedings{cordts2016cityscapes,
  title={The cityscapes dataset for semantic urban scene understanding},
  author={Cordts, Marius and Omran, Mohamed and Ramos, Sebastian and Rehfeld, Timo and Enzweiler, Markus and Benenson, Rodrigo and Franke, Uwe and Roth, Stefan and Schiele, Bernt},
  booktitle={Proc. CVPR},
  pages={3213--3223},
  year={2016}
}

@inproceedings{caesar2018coco,
  title={Coco-stuff: Thing and stuff classes in context},
  author={Caesar, Holger and Uijlings, Jasper and Ferrari, Vittorio},
  booktitle={Proc. CVPR},
  pages={1209--1218},
  year={2018}
}

@inproceedings{mottaghi2014role,
  title={The role of context for object detection and semantic segmentation in the wild},
  author={Mottaghi, Roozbeh and Chen, Xianjie and Liu, Xiaobai and Cho, Nam-Gyu and Lee, Seong-Whan and Fidler, Sanja and Urtasun, Raquel and Yuille, Alan},
  booktitle={Proc. CVPR},
  pages={891--898},
  year={2014}
}

@article{everingham2010pascal,
  title={The pascal visual object classes (voc) challenge},
  author={Everingham, Mark and Van Gool, Luc and Williams, Christopher KI and Winn, John and Zisserman, Andrew},
  journal={International Journal of Computer Vision},
  volume={88},
  pages={303--338},
  year={2010},
  publisher={Springer}
}

@inproceedings{gu2021open,
  title={Open-vocabulary Object Detection via Vision and Language Knowledge Distillation},
  author={Gu, Xiuye and Lin, Tsung-Yi and Kuo, Weicheng and Cui, Yin},
  booktitle={Proc. ICLR},
  year={2021}
}

@inproceedings{deng2009imagenet,
  title={Imagenet: A large-scale hierarchical image database},
  author={Deng, Jia and Dong, Wei and Socher, Richard and Li, Li-Jia and Li, Kai and Fei-Fei, Li},
  booktitle={Proc. CVPR},
  pages={248--255},
  year={2009},
}

@inproceedings{bousselham2024grounding,
  title={Grounding everything: Emerging localization properties in vision-language transformers},
  author={Bousselham, Walid and Petersen, Felix and Ferrari, Vittorio and Kuehne, Hilde},
  booktitle={Proc. CVPR},
  pages={3828--3837},
  year={2024}
}

@inproceedings{xing2023rewrite,
  title={Rewrite caption semantics: Bridging semantic gaps for language-supervised semantic segmentation},
  author={Xing, Yun and Kang, Jian and Xiao, Aoran and Nie, Jiahao and Shao, Ling and Lu, Shijian},
  booktitle={Proc. NeurIPS},
  volume={36},
  pages={68798--68809},
  year={2024}
}

@inproceedings{wang2024learn,
  title={Learn to rectify the bias of clip for unsupervised semantic segmentation},
  author={Wang, Jingyun and Kang, Guoliang},
  booktitle={Proc. CVPR},
  pages={4102--4112},
  year={2024}
}

@article{wang2024curriculum,
  title={A curriculum-style self-training approach for source-free semantic segmentation},
  author={Wang, Yuxi and Liang, Jian and Zhang, Zhaoxiang},
  journal={IEEE Transactions on Pattern Analysis and Machine Intelligence},
  volume={46},
  number={12},
  pages={9890--9907},
  year={2024},
  publisher={IEEE}
}

@inproceedings{liang2024realistic,
  title = {Realistic unsupervised CLIP fine-tuning with universal entropy optimization},
  author = {Liang, Jian and Sheng, Lijun and Wang, Zhengbo and He, Ran and Tan, Tieniu},
  booktitle={Proc. ICML},
  pages={29667--29681},
  year={2024}
}

@article{dong2026adapting,
  title={Adapting vision-language models without labels: A comprehensive survey},
  author={Dong, Hao and Sheng, Lijun and Liang, Jian and He, Ran and Chatzi, Eleni and Fink, Olga},
  journal={International Journal of Computer Vision},
  volume={134},
  number={8},
  pages={389},
  year={2026},
  publisher={Springer}
}

\end{document}